# Stemming - The Evolution and Current State with a Focus on Bangla

**Abhijit Paul[1], Mashiat Amin Farin[1], Sharif Md. Abdullah[1]**
Ahmedul Kabir[1]
Zarif Masud[2]
Shebuti Rayana[3]

[1]*Inst. of Information Technology, University of Dhaka, Bangladesh*
[2]*Toronto Metropolitan University, Canada*
[3]*State University of New York, Old Westbury, USA*

## Abstract

Bangla, the seventh most widely spoken language worldwide with 300 million native speakers, faces digital under-representation due to limited resources and lack of annotated datasets. Stemming, a critical preprocessing step in language analysis, is essential for low-resource, highly-inflectional languages like Bangla, because it can reduce the complexity of algorithms and models by significantly reducing the number of words the algorithm needs to consider. This paper conducts a comprehensive survey of stemming approaches, emphasizing the importance of handling morphological variants effectively. While exploring the landscape of Bangla stemming, it becomes evident that there is a significant gap in the existing literature. The paper highlights the discontinuity from previous research and the scarcity of accessible implementations for replication. Furthermore, it critiques the evaluation methodologies, stressing the need for more relevant metrics. In the context of Bangla's rich morphology and diverse dialects, the paper acknowledges the challenges it poses. To address these challenges, the paper suggests directions for Bangla stemmer development. It concludes by advocating for robust Bangla stemmers and continued research in the field to enhance language analysis and processing.

## 1 Introduction

Stemming is the process of condensing a word to its base form, often involving the removal of prefixes and suffixes, and connecting it to a fundamental root, "lemma" [1].

A lemma serves as the base form from which all inflected forms derive. For instance, the words "walk," "walking," "walked," and "walks" share the same lemma, "walk." As a pre-processing step, a stemming algorithm should also exhibit other essential properties.

- It should be lightweight as stemming is almost always used as a preprocessing step [27].

- Stemming need not always reach the lexicographical root; it suffices when morphologically related variants share the same stem, a crucial flexibility for its widespread application in diverse domains [31].

- It must not change the meaning of the original sentence.

Stemming in a highly inflectional language like Bangla, with its numerous word forms stemming from the same roots, poses unique challenges. Despite being one of the world's most widely spoken languages, with nearly 300 million native speakers, Bangla's rich morphology, diverse dialects, and extensive literary history developed over millennia make it a complex linguistic landscape. However, it remains a low-resource language in the digital realm, primarily due to the shortage of annotated, machine-readable datasets and limited resource support.

## 2 Objective

This research seeks to conduct an in-depth survey of Bangla stemming methods. Its main objectives are to comprehend the current landscape of Bangla stemming algorithms, identify challenges, and propose solutions for this low-resource language. Additionally, the study analyzes the trends in stemming research in both Bangla and English, offering a comprehensive perspective. Our contributions in the current study are as follows:

- Defining stemmers in the context of modern NLP.

- Highlighting flaws in evaluating supervised stemming approaches, where accuracy falls short as a metric. Instead, it's crucial to estimate under-stemming and over-stemming degrees.

- How current Bangla stemming works are strangely disconnected from the strides made in FIRE (Forum for Information Retrieval Evaluation) 2008-2013 stemming works [2].
- Trends of stemming research focus between English and Bangla.
- Why Stemming is important for a highly inflectional language like Bangla.
- Future direction to develop a Bangla stemmer.

## 3 Stemming Techniques

We have identified the approaches that are commonly used for stemming tasks from our literature review in Section-4. The approaches are shown in Fig-1.

## 4 Stemming in Bangla

Early attempts, dating back to around 2008, aimed at addressing the renowned challenge of Indian information retrieval using the FIRE 2008 dataset [2]. Consequently, most initial stemming endeavors were geared toward Information Retrieval tasks.

Notably, the period from FIRE 2008-2013 marked significant progress in Bangla stemming. However, it's intriguing that contemporary Bangla stemmer algorithms seldom reference the works from 2008-2013. It's as if they're embarking on the task anew, with little acknowledgment of earlier contributions in the current stemming literature.

### 4.1 FIRE 2008-13, India

FIRE information retrieval contest was a series of annual competitions held from 2008 to 2013 to evaluate the performance of information retrieval systems for Indian languages, including Bangla [2].

The FIRE contest used a variety of corpora and evaluation metrics, including the following:

- **WebIR-Hindi:** A collection of Hindi web documents.
- **ClueWeb09-Hindi:** A collection of Hindi news documents.
- **NDCG:** A measure of the quality of a ranked list of documents.
- **ERR:** A measure of the effectiveness of a retrieval system [2].

Table 1: Stemming Approaches developed during FIRE 2008-13

| Authors | Year | Approach |
| --- | --- | --- |
| Ganguly et al [4] | 2013 | Rule Based |
| Mitra et al [7] | 2012 | Ensemble |
| Banerjee and Pal [5] | 2011 | Statistica |
| L. Dolamic and J. Savoy [6] | 2010 | N-Truncating |
| Mitra et al [7] | 2012 | Ensemble |

The FIRE contest was a valuable resource for researchers and practitioners working on information retrieval for Indian languages. The results of the contest helped to identify the best performing systems and the challenges that still need to be addressed in this area.

Information Retrieval tasks can benefit from stemming and for this reason, this contest resulted in the creation of a lot of complex Bangla stemmers. Table-1 shows some stemming approaches used in FIRE 2008-13.

#### 4.1.1 Review of Fire Literature

According to Dolamic et al., trunc-n indexing or an aggressive stemming technique results in more successful retrieval when compared to alternative word-based or language-independent approaches [6].

An apriori-like algorithm from market basket data analysis was used by S. Pal et al. to develop a statistical stemmer based on frequent pattern mining that improved retrieval performance by 9% over no-stem runs [5].

In the 2012–2013 FIRE competition, Ganguly et al. proposed a rule-based stemmer. The manual rules for eliminating the often occurring plural suffixes from Hindi and Bangla were created using linguistic understanding. Rules were also developed for the removal of Bangla's case markers and classifiers.

One of the best ensemble stemming approaches is Muladhaar [7]. The accuracy of the tool reached up to 96% in internal evaluation on classic literature and contemporary travelogue domains.

### 4.2 Bangladesh 2010-2023

A comprehensive survey on stemming in Bangla is provided. A summary of the survey is shown in Table-2.

Hoque et al. proposed a basic rule-based stem-mer. They used it with a verb dataset, POS tagging rules, and a dictionary of some POS-tagged words, and achieved an accuracy of 93.7%, as shown in Table-3 [8].

Table 2: Bangla Stemming Literature Overview

| Author name | Approach | Evaluation Technique | Evaluation Score |
|---|---|---|---|
| Nesarul and Seddiqui, 2015 [8] | Rule Based | POS Tagging | 93.70% |
| Majumder et al, 2007 [9] | Clustering | Information Retrieval Performance | 49.6 |
| Urmi, Jammy and Ismail, 2016 [10] | World Similarity, N-Gram based, Contextual Similarity, Clustering | Accuracy | 40.18% |
| Das and Bandyopadhyay, 2010 [11] | MED based (Insertion at beginning, Deletion at end), Clustering | Accuracy | 74.60% |
| Seddiqui, Maruf, andChy, 2016 [12] | Recursive Suffix Stripping | Information Retrieval | 92% |
| Das and Mitra, 2011 [13] | Rule Based | Information Retrieval | 47.50% |
| Sarkar and Bandyopadhyay, 2008 [23] | Rule Based | Accuracy, Precision, Recall | rt1: 98.8%, rt2: 98.7%, rt3: 99.9% |
| Islam, Uddin and Khan, 2007 [14] | Suffix Stripping | Spelling Checker | single error acc: 90.8% multi error acc ∼67% |
| Mahmud et al, 2014 [15] | Rule Based | Accuracy | verb: 83% noun: 88% |
| Salim, Ahmed and Hasan, 2019 [17] | Porter-like Structure, Rule-based approach but few rules. | From যোগাযোগ (রবীন্দ্রনাথ ঠাকুর), 174 words were collected for evaluation. | 95% |
| Sadia, Rahman and Seddiqui, 2019 [18] | Character-based, N-gram, Affinity Propagation, K-means Clustering, Similarity Coefficient | Using affinity propagation, they experimented with a dataset of 47k words and identified 10k unique words from it. | 87% |
| Ahmed et al, 2021 | a concise dataset with suffixes and matching stems/morphological words. | - | - |
| Islam, Uddin and Khan, 2010 [14] | Rule Based, separate rule-set for verb, noun and adjective. | A corpus of 13k size where 600 unique words were present. | 90.80% |
| Dolamic and Savoy, 2008 [33] | 4gram and Light Stemmer | Accuracy | light: 41.3 4gram: 40.7 |
| Paik and Parui, 2008 [24] | TERRIER | Accuracy | 42.30% |

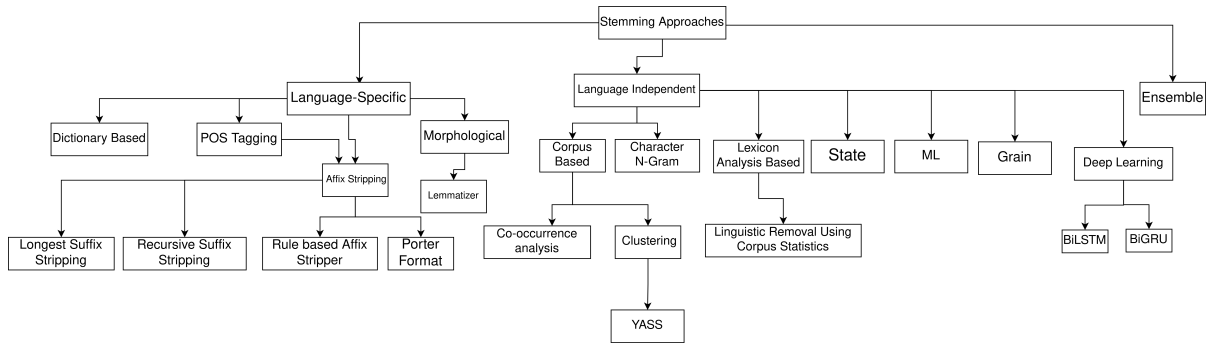

Figure 1: Existing Stemming Techniques

Table 3: POS Tagger with Stemmer [8]

| Total Words | Experiment Type | Detected Words | Accuracy |
|---|---|---|---|
| 8155 | Dictionary + Stemmer | 3842 | 47.10% |
| | Dictionary + Stemmer + Verb Dataset | 5169 | 63.40% |
| | Dictionary + Stemmer + Verb Dataset + Rules | 7637 | 93.70% |

Majumder et al. introduced YASS (Yet Another Suffix Stripper), a clustering-based strategy. They evaluated this approach alongside Porter's and Lovin's stemming methods using 200 queries on the AP and WSJ subsets of the Tipster dataset. YASS exhibited performance akin to Porter's and Lovin's stemmers, with similar results in terms of both average precision and the total count of relevant documents retrieved [9]. . Urmi et al. employed a corpus-based unsupervised approach with a 6-gram language model on Bangla data, achieving 40.18% accuracy [10].

In Das et al. [11], K-means clustering was applied to the IL-ILMT dataset, resulting in an accuracy of 74.6%. This marks one of the few clustering implementations in Bangla stemming.

Seddiqui et al. introduced a recursive suffix stripper for Bangla stemming. Their method involves stemming inflectional words in various ways using the recursive suffix stripping algorithm.

They collected 0.78 million words from the Prothom Alo newspaper and achieved an overall accuracy of 92% on this dataset [12]. .

Das and Mitra presented a rule-based stemming algorithm for inflectional and derivational words in Bangla. Using the FIRE 2010 Bangla test collection with 50 queries, they reported a MAP value of 0.4748 in their experiment [13].

Islam et al. developed a lightweight stemmer for a Bangla spellchecker, utilizing a lexicon of 600 root words and a list of 100 suffixes. They tested it with 13,000 words, achieving a single error accuracy of 90.8% and a multi-error accuracy of approximately 67% [14]. .

In [15], the authors presented a rule-based approach, notable as the first stemmer not dependent on consistent dictionary searches for validation. They distinguished verb and noun inflections separately, achieving 88% accuracy for verbs and 83% for nouns.

In [17], the authors devised a rule-based stemmer by drawing inspiration from grammatical rules found in the class 9-10 Bangla Grammar book. They evaluated their algorithm using 174 words gathered from যোগাযোগ (রবীন্দ্রনাথ ঠাকুর) and reported a high accuracy of 95% through manual inspection. This work is noteworthy for two reasons: it assessed stemming accuracy through the stemming task itself and showcased how inspiration from grammatical rules can significantly enhance accuracy.

In [18], the authors introduced a clustering approach to stemming. They employed the dice coefficient, an affiliation measure, to cluster sets of words based on their character structure. The smallest word within a cluster was considered as the stem. They also explored Affinity Propagation clustering algorithms with coefficient and median similarity, reporting an accuracy of approximately

### 4.3 Other Related Works

While not directly focusing on stemming tasks, these research works can greatly benefit the making of a Bangla stemming algorithm.

In [8], Hoque et al. introduced a POS-tagging system, offering a dataset of 45,000 words with their default tags and a patterned verb dataset. This dataset enables stemming researchers to employ a dictionary-based approach for stemming verbs.

Dash, in [19], presented an automated method for processing pronouns in a Bangla text corpus comprising 3.5 million words. The author conducted a corpus-based analysis of Bangla pronouns and devised an innovative approach for their analysis. This led to the development of a system capable of identifying and analyzing Bangla pronouns in corpus data.

In [20], Rabbi et al. proposed a novel technique for suffix stripping in Bangla grammar pattern recognition. They also described a parsing method for Bangla grammar recognition, involving a shift-reduce parser that constructs a parse table based on the LR (Left-to-right, Rightmost derivation) strategy. This strategy processes input strings from left to right, performing rightmost derivations of the CFG (Context-Free Grammar) to recognize and build the parse tree for the given input.

### 4.4 Resources for Bangla Stemming

Sakhawat Hossain et al. [21] developed a dataset of suffixes and their relevant stems/morphological words. The corpus has 1000 words.

Amitava Das et al. [11] used Indian Languages Machine Translation System's (IL-ILMT-2) gold standard Morphological dataset for evaluation of their approach.

## 5 Evolution Trends

### 5.1 General Trends among Languages

Fig-2 highlights the trends in stemming research in English vs Bangla.

In our opinion, the evolution of stemming can be divided into the following five steps:

- Initial Work: Foundational research often marks the inception of a field, typically emerging as a subtask within a larger context. For instance, in Information Retrieval tasks, there was a growing need for a more sophisticated stemmer beyond simple truncation, which led to the development of stemming algorithms.

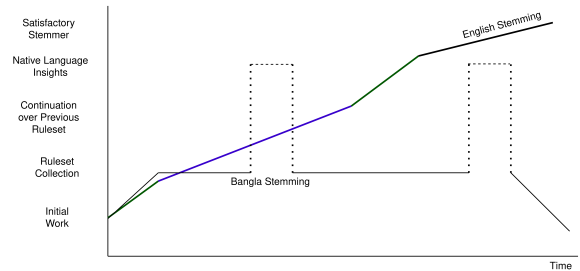

Figure 2: Trends of Stemming Research Focus among Languages

- Ruleset Collection: Stemming tasks started in english through ruleset collection. Firstly, there is Lovin's rule-based algorithm [29] for the English language. This algorithm lists a set of suffixes and what they should be replaced with. It also includes some context-sensitive rules that apply to specific suffixes. It had

  – 294 Suffixes, 29 context sensitive rules for each of them.
  – 34 recoding rules for stem. (Absorption → absorp → absorb).

- Continuation Over Previous Ruleset: In English, Dawson's stemmer [30] extended Lovin's approach with a comprehensive list of about 1200 suffixes, improving upon the original stemmer.

- Native Language Insight: In English, stemming rulesets can be systemized using insights from the native language, such as considering the minimum average stem length and syllable lengths. For example, Porter [32] introduced a more sophisticated approach by using a measure called the "m-value," which is based on the number of consonant-vowel-consonant strings remaining after suffix removal. This approach is derived from earlier work by Dolby and Resnikoff (1964) [39].

  For Bangla, native language insight can be minimum stemming length which comes around 3 or 4 letters. More discussion on it in Section 6.

- Satisfactory Stemmer: A satisfactory stemmer, as previously defined, is essentially a lightweight tool that notably reduces

over-stemming, even if it increases under-stemming to some extent. The primary criterion for satisfaction lies in its ability to serve as a universal preprocessing step for text in various NLP tasks without compromising the semantic content of the text. Porter stemmer serves as a prime example of such a tool.

### 5.2 Issues with Bangla Stemming Evolution So Far

Bangla stemming research has not yet developed any satisfactory stemmer. The following issues have been identified as responsible for that.

- **Bangla Dataset:** Bangla stemming datasets are limited. Section 4.4 mentions two datasets from 2016 and 2021, but their sources are now inaccessible. Given this scarcity, evaluations and training are often incorporated into larger tasks like information retrieval, spelling correction, or POS-tagging systems.

- **Lack of Continuation over Previous Ruleset:** Most Bangla stemming research starts from scratch without building upon prior rule sets.

  The progress made in creating Bangla stemmers during the FIRE Information Retrieval contests from 2008 to 2013 seems to have been largely overlooked in current Bangla stemming literature.

- **Bangla Stemming Evaluation:** All Bangla stemmers used accuracy as their evaluation metric and none of them used over-stemming, or under-stemming, or similar intrinsic evaluation method for stemmers. Accuracy is a poor estimate for evaluating stemming tasks because it is not necessarily inaccurate if a word is under-stemmed. More discussion on it in section-6.3.

## 6 Suggestions and Future Directions

In this section, Some issues that are pertinent to most research works in stemming are discussed, along with what can be done to build a proper stemmer for Bangla.

### 6.1 Why stemming is important for Bangla?

According to Porter, highly inflectional languages benefit greatly from stemming [32]. Bangla is a highly inflectional language, and as such, stemming could vastly improve performance in various Bangla NLP tasks.

Table 4: Morphological Variants of each POS [22]

| Parts of Speech | Average Morphological Variations |
|---|---|
| বিশেষ্য (Noun) | 36 |
| সর্বনাম (Pronoun) | 24 |
| ক্রিয়া (Verb) | 168 |

In Bangla, there are 5 parts of speech. They are: বিশেষ্য (Noun), বিশেষণ(Adjective, Adverb), সর্বনাম (Pronoun), অব্যয় (Conjunction, Preposition) and ক্রিয়া (Verb).

In Bangla, verbs exhibit extensive inflections, with around 200 morphological variants for a single verb due to the addition of various suffixes. Traditional suffix stripping methods are less effective for Bangla verb stemming. Other parts of speech in Bangla have fewer morphological variations. For more details, refer to Table-4 [22].

Effective Bangla word stemming can significantly reduce the complexity of NLP tasks, especially in handling large vocabularies. This simplification results in reduced computation time and model complexity. It offers substantial advantages for various tasks, including word embeddings, Information Retrieval, data mining, spelling corrections, and more.

### 6.2 Why grammatical rules falls-short for stemming

Many NLP researchers believe that collecting grammatical rules alone is adequate for stemming tasks. However, this notion is challenged in the following section. Stemming aims to unify morphologically related words into a single form, rather than identifying the lexicographical root (as discussed in Section-1). Therefore, relying solely on grammatical rules may not be suitable for the task.

Let's consider the words *jumper* and *jumped*. A stemmer rule such as the one provided below would cause over-stemming:

- If the word ends in "er", replace it with null.
- If the word ends in "ed", replace it with null.

The above stemming algorithm may reduce both the words to the stem "jump", which would imply that both words mean the same things, but this is clearly wrong. As jumper is a piece of clothing and jumping is an activity.

Notably, grammatical rules contain a lot of exceptions. Hence there is a serious chance of over-stemming if only grammatical rules are used for stemming.

Then a question may arise - how to formulate the rules for a rule-based stemmer without following the traditional grammatical rules? The answer lies in Lovin's approach. More discussion on this is presented in Section 6.

### 6.3 Dangers of Over-Stemming

The primary objective of stemming is to unify morphologically related words into the same word-form. However, over-stemming can lead to incorrect stems, altering the semantics of the original word. This problem, known as semantic erosion, undermines the purpose of stemming as a preprocessing step. Consequently, a reliable stemming algorithm should minimize over-stemming. Hence, accuracy is an insufficient metric for evaluating stemming tasks.

### 6.4 Suggestions for the development of Standard Stemming Algorithm

The following suggestions are made as a general direction for the development of standard stemming algorithm.

- Using Lovins-like approach: Lovins used an iterative approach for rule development, adjusting and refining rules based on evaluation results, which proved more effective than collecting grammatical rules alone [29].

- Consider semantics for better results: Clustering and rule-based approaches can significantly benefit from semantic information since morphologically related words often appear in similar contexts.

- Consider trade-offs between computation cost and performance: Surprisingly, there are no deep learning-based stemming approaches. The reason is obvious: stemming has traditionally been a lightweight preprocessing task. Introducing high-computational-cost approaches may improve model performance but wouldn't serve the purpose of an efficient stemmer.

### 6.5 Stemming Datasets

Datasets are vital for any supervised or rule-based approach to stemming. Fortunately, stemming datasets can be easily found from Bangla Dictionary as each word has their corresponding roots. Words with the same root can be clustered, and this can be used to estimate over-stemming, under-stemming degree, accuracy, etc.

### 6.6 A Basic Framework for Stemming in Bangla

Systematic, focused research is needed to develop satisfactory stemmers for Bangla. In this regard, a framework is proposed in this paper, as shown in Figure 3.

In the context of Bangla language stemming, a character can be defined as one of the following, as detailed in [23]:

- An independent basic consonant or vowel
- A basic consonant along with a succeeding dependent vowel.
- A compound character with or without a succeeding dependent vowel.

There is another measure however and that is Orthographic syllable, the basic orthographic unit of Bangla [24].

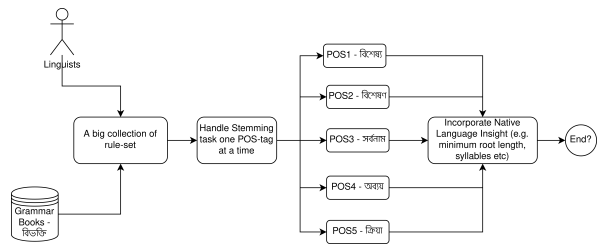

Figure 3: Recommended Direction of Stemming Research for Bangla

Secondly, suffix sets need to be identified for nouns, pronouns, and verbs. Each part-of-speech (POS) usually has its own independent set of suffixes. Next, it is necessary to find how many inflections are actually a result of multiple suffixes and for which parts-of-speech (POS) they occur.

Thirdly, rules need to be made for Bangla language. The rules do not necessarily need to be based on grammar. However, here are some observations from the work of Sandipan Sarkar et al. [23].

It's important to note that Bangla has a wide range of verb roots, approximately 1500 or more,

categorized into 20 different types [15]. This diversity allows for the use of a dictionary-based approach for verbs.

### 6.7 About Snowball Framework

The Snowball framework is a powerful tool for crafting stemming algorithms tailored for Information Retrieval purposes. Originally developed by Martin Porter, it's now a community-driven project after his retirement in 2014.

Snowball offers a compiler that translates programs written in its scripting language into source code compatible with multiple programming languages, including Ada, ISO C, Go, Java, Javascript, Object Pascal, Python, and Rust.

Using the Snowball framework is recommended for developing stemming algorithms, as it enables researchers to incorporate previous rulesets easily into their work and ensures consistent reproducibility [25].

### 6.8 Evaluation

Stemming works are often assessed in Information Retrieval, POS tagging, and spelling correction. Intrinsic evaluation is essential, focusing on over-stemming and under-stemming. Chris D. Paice's [26] work provides a valuable method for estimating these factors. It considers over-stemming and under-stemming count to estimate the performance of stemming algorithm.

## 7 Conclusion

More specifically, This research underscores the importance of stemming in highly inflectional languages like Bangla. It provides a thorough survey of stemming methods across languages, emphasizing the need for better stemmers to handle morphological variations. The study highlights the importance of using under-stemming and over-stemming as evaluation metrics for supervised stemming.

It addresses the unique challenges faced by Bangla due to its rich morphology and limited digital resources. The paper suggests directions for a Bangla stemmer and encourages further research in this area to enhance language analysis and processing for Bangla.

## 8 Ethical Considerations and Limitations

The study was conducted on publicly available data sources. No ethical dilemma were identified. However, this study has a limitations. The sample size was relatively small, which limits the generalizability of the findings.

Despite these limitations, this study provides valuable insights into the current state of Bangla stemming. The findings suggest that new stemming research works should build up on the previous works. Further research with larger sample sizes, longitudinal designs, and objective measures is needed to confirm these findings.